\pdfoutput=1
\documentclass[9pt,twocolumn]{extarticle}
\usepackage[a4paper, textwidth=180mm, textheight=250mm, top=18mm,
  columnsep=6mm, footskip=8mm]{geometry}
\PassOptionsToPackage{table}{xcolor}

\usepackage[T1]{fontenc}
\usepackage[utf8]{inputenc}
\usepackage[english]{babel}
\usepackage{amsmath,amssymb,amsthm}
\usepackage{newtxtext}
\usepackage{newtxmath}
\ifdefined\XeTeXversion
  \usepackage{fontspec}
\else
  \usepackage[scaled=0.92]{helvet}
\fi
\usepackage[scaled=0.95]{zlmtt}
\usepackage{microtype}
\usepackage[dvipsnames]{xcolor}
\usepackage{graphicx}
\usepackage{booktabs}
\usepackage{colortbl}
\usepackage{tabularx}
\usepackage{longtable}
\usepackage{array}
\usepackage{enumitem}
\usepackage{url}
\usepackage{xurl}  %
\usepackage[super,sort&compress,comma]{natbib}
\usepackage{caption}
\usepackage{titlesec}
\usepackage{fancyhdr}
\usepackage[hidelinks]{hyperref}
\hypersetup{colorlinks=true, linkcolor=black, citecolor=black, urlcolor={blue!70!black}}

\definecolor{methodcolor}{HTML}{B03A2E}

\titleformat{\section}{\sffamily\bfseries\fontsize{11}{13}\selectfont}{}{0pt}{}
\titlespacing*{\section}{0pt}{10pt plus 2pt minus 2pt}{3pt}
\titleformat{\subsection}{\sffamily\bfseries\fontsize{9.5}{11.5}\selectfont}{}{0pt}{}
\titlespacing*{\subsection}{0pt}{8pt plus 2pt minus 2pt}{2pt}
\titleformat{\paragraph}[runin]{\bfseries}{}{0pt}{}
\titlespacing*{\paragraph}{0pt}{3pt plus 1pt}{0.5em}

\DeclareCaptionLabelSeparator{nature}{\,\textbar\ }
\addto\captionsenglish{}  %

\newcommand{\firstpagenote}[1]{{\let\thefootnote\relax\footnotetext{#1}}}

\setlist{nosep, leftmargin=*}

\usepackage{marvosym}  %
\usepackage{fontawesome5}  %
\usepackage{mathtools}
\usepackage{bm}
\usepackage{algorithm}
\usepackage{algorithmic}   %
\usepackage{multirow}
\usepackage{float}
\usepackage{placeins}   %
\usepackage{tikz}
\usetikzlibrary{arrows.meta,positioning,calc,fit}
\usepackage{cleveref}

\theoremstyle{plain}

\theoremstyle{definition}

\theoremstyle{remark}

\newcommand{\std}[1]{{\scriptsize\,$\pm$#1}}

\crefname{figure}{Fig.}{Figs.}\Crefname{figure}{Fig.}{Figs.}
\crefname{table}{Table}{Tables}\Crefname{table}{Table}{Tables}
\crefname{equation}{equation}{equations}\Crefname{equation}{Equation}{Equations}
\creflabelformat{equation}{(#2#1#3)}
\crefname{algorithm}{Algorithm}{Algorithms}\Crefname{algorithm}{Algorithm}{Algorithms}

\usepackage{multicol}
\usepackage{verbatim}       %
\usepackage{wrapfig}
\usepackage{stfloats}   %
\usepackage{adjustbox}
\usepackage{makecell}
\usepackage{subcaption}
\usepackage{pifont}
\usepackage{siunitx}
\usepackage{listings}
\usepackage[most]{tcolorbox}  %

\usepackage{amsmath,amsfonts,bm}

\def\eqref#1{equation~\ref{#1}}

\def\1{\bm{1}}

\DeclareMathAlphabet{\mathsfit}{\encodingdefault}{\sfdefault}{m}{sl}
\SetMathAlphabet{\mathsfit}{bold}{\encodingdefault}{\sfdefault}{bx}{n}

\definecolor{spblue}{HTML}{00b5ea}
\definecolor{spred}{HTML}{E74C3C}
\definecolor{spgreen}{HTML}{2ECC71}
\definecolor{lightblue}{rgb}{0.88, 0.95, 0.98}
\definecolor{lightpink}{rgb}{0.98, 0.88, 0.95}
\definecolor{superlightred}{rgb}{0.99, 0.92, 0.92}

\lstdefinestyle{demo}{
  backgroundcolor=\color{white},
  basicstyle=\ttfamily\footnotesize,
  breaklines=true,
  captionpos=b,
  keepspaces=true,
  numbers=left,
  numbersep=15pt,
  showspaces=false,
  showstringspaces=false,
  showtabs=false,
  tabsize=2,
  frame=none
}

\newcommand{\icon}[2][0.4cm]{\begin{minipage}{#1}\includegraphics[width=#1]{#2}\end{minipage}}
\newcommand{\gpt}[0]{\icon{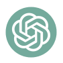}}
\newcommand{\claude}[0]{\icon{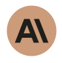}}
\newcommand{\llama}[0]{\icon{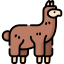}}
\newcommand{\deepseek}[0]{\icon{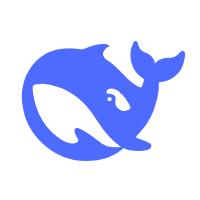}}
\newcommand{\qwen}[0]{\icon{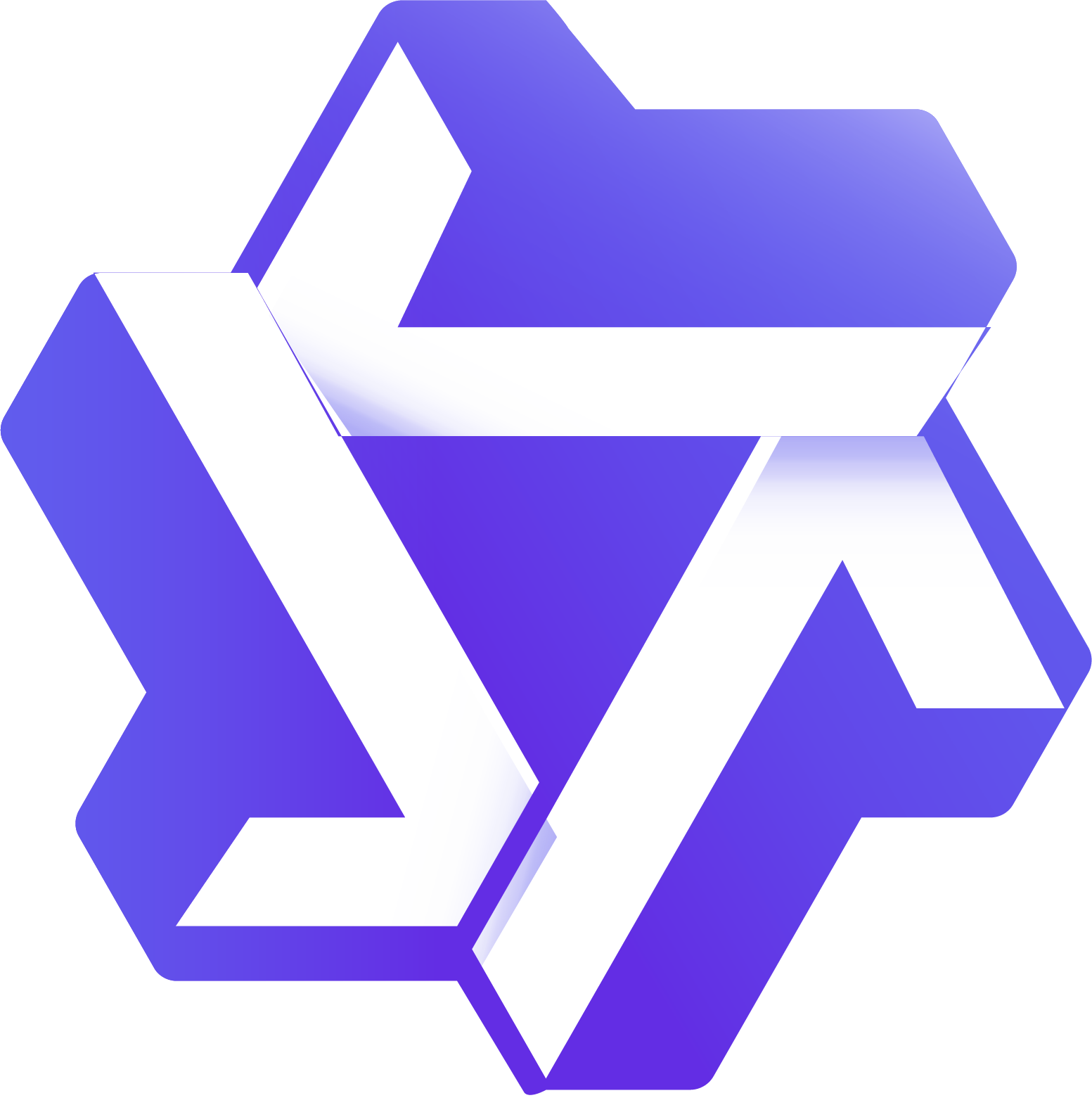}}
\newcommand{\grok}[0]{\icon{}}
\newcommand{\gemini}[0]{\icon{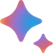}}

\newcommand{\mt}{\texttt{ScienceEval}}
\newcommand{\st}{\texttt{ScienceNarrative}}

\makeatletter
\newcommand{\xr@lookup}[2]{%
  \@ifundefined{xr@#1@#2}%
    {\textbf{??}\@latex@warning{Cross-document reference `#2' undefined}}%
    {\@nameuse{xr@#1@#2}}}
\newcommand{\sref}[1]{\xr@lookup{S}{#1}}
\newcommand{\mref}[1]{\xr@lookup{M}{#1}}
\newcommand{\xr@def}[3]{\expandafter\gdef\csname xr@#1@#2\endcsname{#3}}
\makeatother

\input{supp_labels}
\definecolor{githubblack}{HTML}{24292E}
\newcommand{\aff}[1]{\raisebox{0.6ex}{\fontsize{6.3}{7}\selectfont #1}}
\newcommand{\papertitle}{{Roundtable Policy}: Confidence-Weighted-Consensus Aggregation Improves Multi-Agent-System Reasoning}
\newcommand{\runningtitle}{Yao et al.\ \textbar\ Roundtable Policy}
\newcommand{\paperauthors}{%
  Yu Yao\aff{1,\dag,\Letter}, Jiayi Dong\aff{3,\dag}, Zhenting Qi\aff{2}, Tianyu Wu\aff{2}, Chengxi Zhang\aff{2},
  Himabindu Lakkaraju\aff{2}, Yang Yang\aff{3}, Shuiwang Ji\aff{4}, Yilun Du\aff{2}, Ju Li\aff{1,\Letter}}
\newcommand{\paperaffiliations}{%
  \aff{1}MIT. \aff{2}Harvard. \aff{3}UCLA. \aff{4}TAMU.
  \aff{\dag}Equal contribution. \aff{\Letter}Correspondence to yu\_yao@mit.edu and liju@mit.edu.}
\newcommand{\papersidebar}{%
  \textbf{Correspondence}\\ yu\_yao@mit.edu\\ liju@mit.edu\\[6pt]
  \textbf{Code and Data}\\
  \href{https://github.com/AI4SciDisc/roundtable}
  {\textcolor{githubblack}{\faGithub} Roundtable/Code}\\[6pt]
  \href{https://huggingface.co/datasets/jiayijdong/Roundtable}
  {\raisebox{-1.5pt}{\hspace{-1pt}\includegraphics[height=9pt]{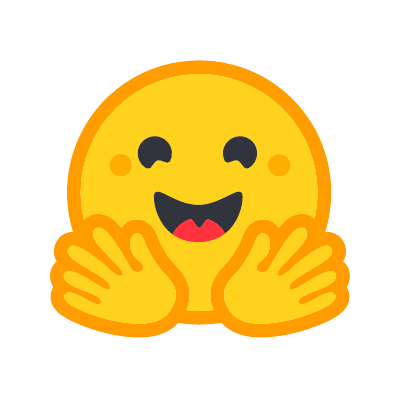}} Roundtable/Data}\\[6pt]
  \textbf{Supplementary Information}\\
  \href{https://github.com/AI4SciDisc/roundtable/tree/main/supplementary}
  {\textcolor{githubblack}{\faFilePdf} Roundtable/SI}\\[6pt]
}

\graphicspath{{figures/}}
\fancypagestyle{plain}{\fancyhf{}\fancyfoot[R]{\sffamily\footnotesize\bfseries\thepage}}

\renewenvironment{abstract}{}{}

\begin{document}

\twocolumn[{%
\begin{flushleft}
\includegraphics[height=14pt]{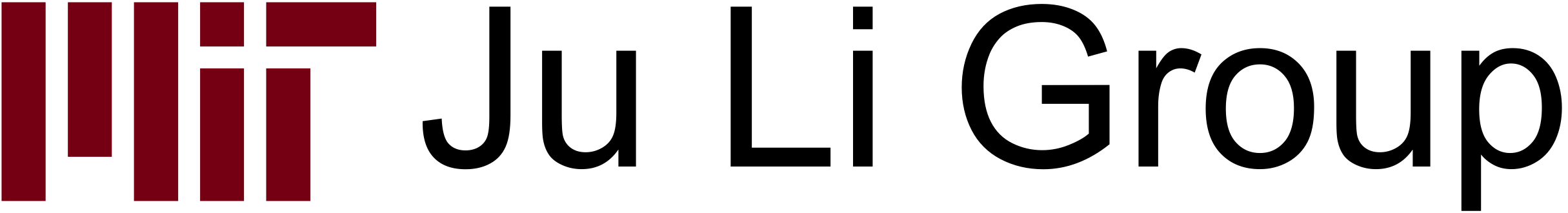}\par
\vspace{3pt}\hrule height 0.6pt\vspace{8pt}
{\sffamily\bfseries\fontsize{21}{24}\selectfont \papertitle\par}
\vspace{8pt}\hrule height 0.4pt\vspace{8pt}
\begin{minipage}[t]{0.24\textwidth}
\sffamily\footnotesize\raggedright \papersidebar
\end{minipage}\hfill
\begin{minipage}[t]{0.73\textwidth}
{\sffamily\bfseries\fontsize{9.5}{12}\selectfont\raggedright \paperauthors\par}
\vspace{9pt}
{\fontsize{10.5}{13.6}\selectfont

\begin{abstract}
Multi-agent systems have demonstrated exceptional performance in downstream tasks beyond diverse single agent baselines. A growing body of work has explored ways to improve their reasoning and collaboration, from vote, debate, to complex interaction protocols. However, it still remains opaque why specific choice would be preferred in multi-agent systems. Inspired by the decision-making mechanism of democratic committees and \emph{The Society of Mind}, we introduce \textsc{Roundtable Policy}, an inference-time reasoning framework for multi-agent systems that performs inference through the weighted consensus of multiple LLMs. Through extensive experiments, we demonstrate its that this approach significantly enhances reasoning in complex heterogeneous scientific tasks. \textsc{Roundtable Policy} emphasizes structured and interpretable inference rather than opaque convergence, while requires only black-box access and uniform procedures, making it broadly applicable to diverse multi-agent systems. 
\end{abstract}

\par}
\end{minipage}
\end{flushleft}
\vspace{10pt}
}]

\firstpagenote{\footnotesize\paperaffiliations}

\begin{figure*}[t]
\centering
\includegraphics[width=\textwidth]{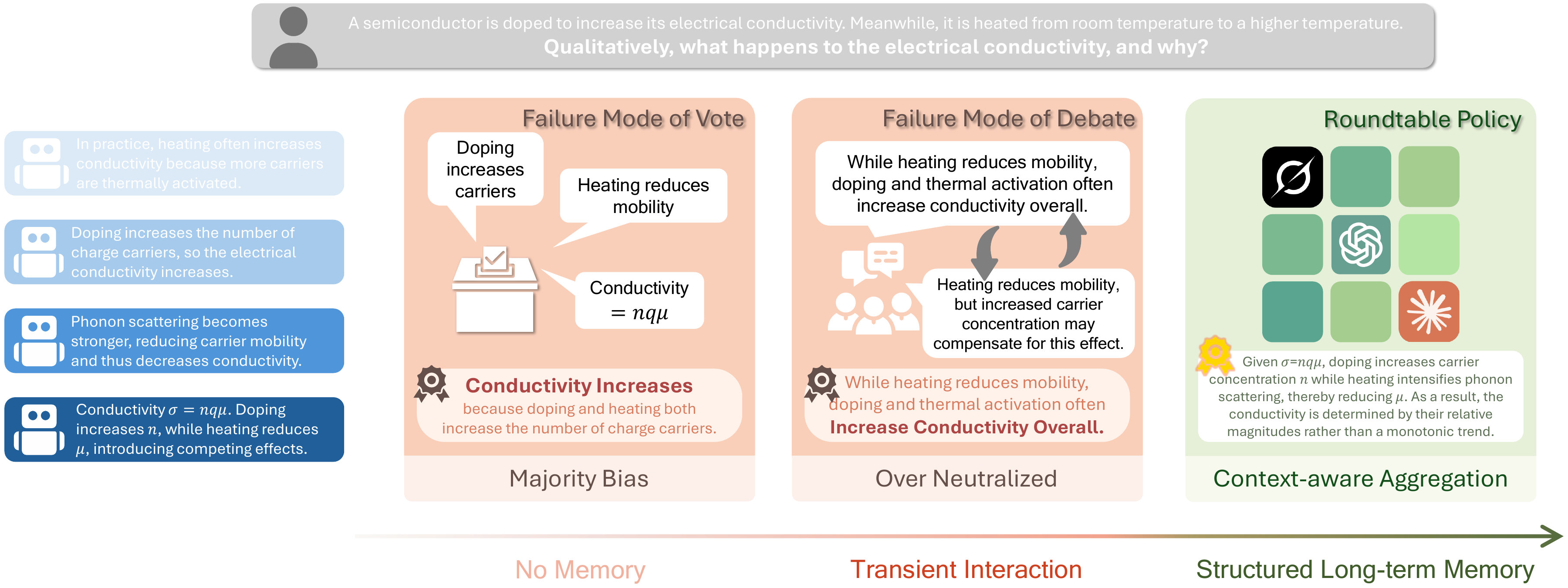}
\caption{\label{fig:failure_modes}
\footnotesize{\textbf{Motivation for \textsc{Roundtable Policy}.} Qualitative illustration of the limitations of existing multi-agent systems and motivation for \textsc{Roundtable Policy}.
\emph{Left:} Voting-based aggregation lacks memory and treats all agents equally, leading to majority bias when partial but confident opinions dominate.
\emph{Middle:} Debate-based interaction relies on transient conversational dynamics and often converges to rhetorically balanced statements.
\emph{Right:} \textsc{Roundtable Policy} introduces a structured, long-term memory of agents' reliability and multi-agent consensus, producing coherent reasoning.}}
\end{figure*}

\noindent Scientific reasoning often requires synthesizing multiple, partially reliable perspectives. This process needs not only predicting a single correct answer, but also forming a coherent and reliable consensus across partially correct, incomplete, or even conflicting explanations. Large language models (LLMs) have shown remarkable capabilities in generating informative responses, motivating their use in scientific discovery and reasoning tasks. However, when applied to answering complex scientific problems, individual models may hallucinate facts, make brittle logical jumps, or emphasize different aspects of a problem. These behaviors reflect not only limitations of individual models, but also a broader challenge: scientific reasoning is not fully captured by isolated, single-shot predictions, but instead requires mechanisms for aggregating multiple uncertain and heterogeneous reasoning paths. For instance, to write a scientific proposal must not only provide factually correct content, but also organize it into a structured and persuasive narrative. Addressing these requirements calls for mechanisms that can reconcile multiple reasoning paths.

A natural motivation is therefore to leverage multi-agent systems~\citep{minsky1988society}, collecting their complementary strengths via diverse reasoning trajectories. Prior work has explored from majority~\citep{wang2022self}, debate~\citep{10.5555/3692070.3692537, liang-etal-2024-encouraging}, to other ensemble approaches~\citep{chen-etal-2024-reconcile,wang2025mixtureofagents}. While effective in many cases, these approaches exhibit inherent limitations: majority vote, treat all models equally, overlooking that different models excel in different domains; Debate-style methods, while sometimes converging to better outputs, do not reliably outperform other prompting strategies~\citep{smit2024mad} and remain inherently opaque: it is unclear why a particular answer was selected, or hard to quantify which models were more trustworthy in the process. These observations all point to a central question:

\begin{center}
\vspace{0pt}
\textbf{\emph{how can we effectively combine complementary strengths across language agents to solve complex scientific tasks?}}
\vspace{0pt}
\end{center}

\begin{figure*}
\centering
\includegraphics[width=\textwidth]{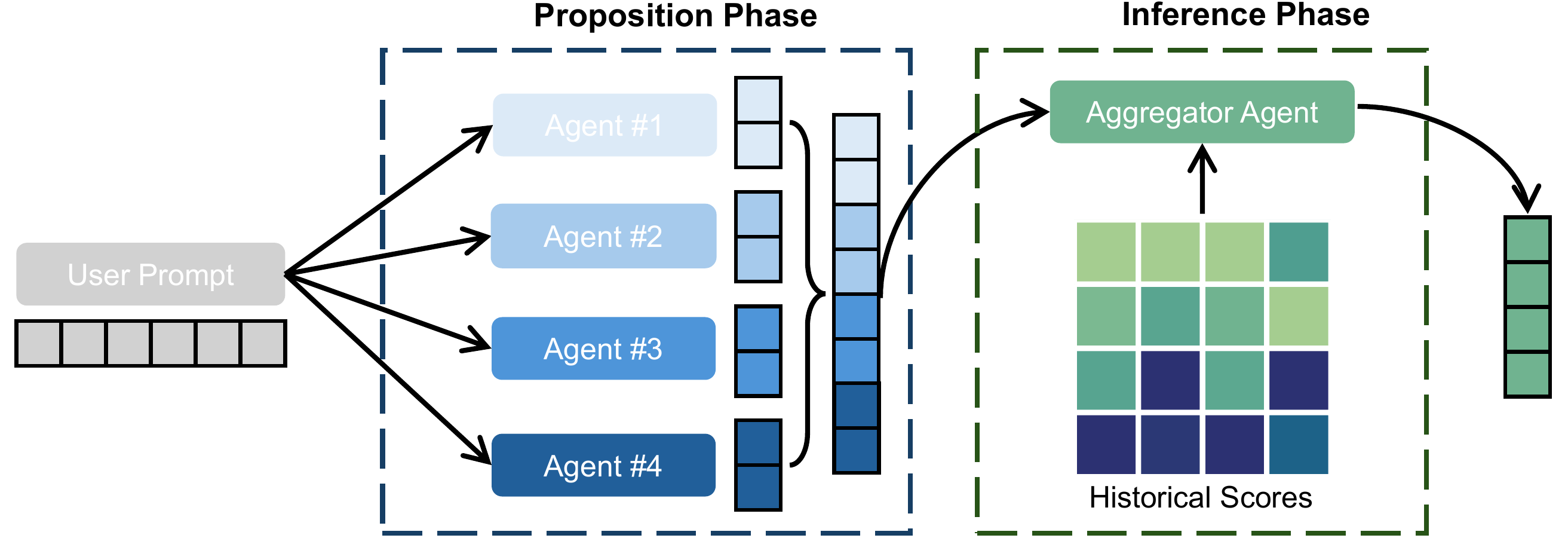}
\caption{\label{fig:abstract_demo}
{\footnotesize \textbf{\textsc{Roundtable Policy}} is an inference-time reasoning framework without retraining or finetuning.}
}

\end{figure*}

We formulate multi-agent reasoning as a problem of consensus formation across heterogeneous reasoning trajectories and instantiate this principle with \textsc{Roundtable Policy}, an inference-time reasoning framework that explicitly models agent reliability with uncertainty across task-specific dimensions (Fig.~\ref{fig:abstract_demo}), producing auditable signals to make inference more transparent and interpretable without modifying base models. Weighting fallible judges by their estimated reliability is a classical idea in statistics~\citep{dawid1979maximum}; here reliability is measured from graded history rather than from the self-reported confidence of LLMs, which is often miscalibrated~\citep{xiong2024uncertainty}. In our framework, diverse agents first propose candidate responses, which are evaluated and recorded across tasks. Then these historical evaluations are distilled into a confidence-weight table that quantifies reliability and uncertainty. During inference, this structured memory guides aggregation, enabling the system to integrate diverse perspectives into a coherent consensus while remaining transparent about the formation of trust.

This perspective also motivates how scientific reasoning systems should be evaluated. Current evaluation has largely relied on benchmarks with isolated, singe-domain tasks. Representative examples such as GSM8K~\citep{cobbe2021gsm8k}, PubMedQA~\citep{jin-etal-2019-pubmedqa}, and AlphaGeometry~\citep{alphageometry2024} have been instrumental in assessing domain-specific knowledge and pointwise accuracy. However, these benchmarks typically assume short-context, single-problem settings, and therefore fail to capture key characteristics of scientific contexts, which often require integrating knowledge across domains and maintaining logical consistency over long contexts. For example, writing a scientific proposal involves synthesizing background knowledge, justifying methodologies, and keeping long-context consistency, rather than solving an isolated arithmetic problem like \emph{4+23*6+24-24*12} (GSM8K). This mismatch becomes sharper in multi-agent systems, which essentially requires aggregating complementary strengths.

To address this gap, we design two complementary benchmarks that capture distinct but essential aspects of scientific reasoning: \textbf{\mt} and \textbf{\st}. \mt\ evaluates cross-domain reasoning by jointly querying agents on heterogeneous problems spanning geoscience~\citep{geobench}, biology~\citep{jin-etal-2019-pubmedqa}, mathematics~\citep{cobbe2021gsm8k}, and multi-faceted physical science (Supplementary Note~\sref{app:bench}). \st\ focuses on long-context reasoning consistency by requiring agents to generate structured scientific proposals under contextual templates. Across both benchmarks, \textsc{Roundtable Policy} consistently improves factual reliability and narrative coherence, achieving average gains of \textbf{13.01\% in \mt} and \textbf{11.04\% in \st} over diverse baseline agents.

\paragraph{Contributions.} (1) We formulate multi-agent reasoning as a problem of consensus formation across heterogeneous reasoning trajectories, and instantiate this perspective with \textsc{Roundtable Policy} and demonstrated its efficacy through extensive experiments.
(2) We introduce two open-ended text generation benchmarks: \mt\ and \st, complementary with domain-authentic testbeds for complex scientific reasoning.

\section{Methodology}

\textbf{Design Intuitions.} In multi-agent reasoning, agents often differ in expertise and reliability across tasks. Effective collaboration therefore requires inference-time mechanisms that can account for these differences, rather than treating all agents or reasoning paths as equally trustworthy. A useful intuition comes from democratic committees. Members independently evaluate a proposal, but final decisions are rarely made by simple voting. Instead, assessments are weighted by expertise, past reliability, and confidence, leading to decisions that are both robust and interpretable. \textsc{Roundtable Policy} follows this intuition by viewing multi-agent reasoning as a committee-style inference process: agents propose diverse reasoning trajectories, their task-specific reliability is assessed over time, and a final consensus is formed by weighting historical reliability and confidence of each member.

\subsection{Multi-Phase Setting}
To realize this intuition, we structure multi-agent reasoning into multiple phases that explicitly separate proposal, and consensus formation. Notations follow Supplementary Note~\sref{app:notation}.

\definecolor{codeblue}{rgb}{0.25,0.5,0.5}
\definecolor{codekw}{rgb}{0.85, 0.18, 0.50}

\definecolor{codesign}{RGB}{0, 0, 255}
\definecolor{codefunc}{rgb}{0.85, 0.18, 0.50}

\lstdefinelanguage{PythonFuncColor}{
  language=Python,
  keywordstyle=\color{blue}\bfseries,
  commentstyle=\color{codeblue},
  stringstyle=\color{orange},
  showstringspaces=false,
  basicstyle=\ttfamily\small,
  mathescape=true,
  literate=
    {*}{{\color{codesign}* }}{1}
    {-}{{\color{codesign}- }}{1}
    {+}{{\color{codesign}+ }}{1}
    {dataloader}{{\color{codefunc}dataloader}}{1}
    {sample_t_r}{{\color{codefunc}sample\_t\_r}}{1}
    {randn}{{\color{codefunc}randn}}{1}
    {randn_like}{{\color{codefunc}randn\_like}}{1}
    {jvp}{{\color{codefunc}jvp}}{1}
    {stopgrad}{{\color{codefunc}stopgrad}}{1}
    {metric}{{\color{codefunc}metric}}{1}
    {sum}{{\color{codefunc}sum}}{3}
    {inference}{{\color{codefunc}inference}}{9}
    {mode}{{\color{codefunc}mode}}{4}
}

\lstset{
  language=PythonFuncColor,
  backgroundcolor=\color{white},
  basicstyle=\fontsize{9pt}{9.9pt}\ttfamily\selectfont,
  columns=fullflexible,
  breaklines=true,
  captionpos=b,
}

\begin{algorithm}[t]
\caption{Inference with Roundtable Policy}
\label{alg:inference}
\begin{lstlisting}
# D = {$\{(q^j, k^j)\}_{j=1}^R$}: task set
# {$\{A_i^P\}_{i=1}^{L_p}$}: players
# {$\{A_l^G\}_{i=1}^{L_g}$}: graders
# {$A^F$}: aggregator
$\vartheta$ = 0
for $j$ = 1, ..., $R$:
    $r^j$ = $\{A_i^P(q^j)\}_{i=1}^{L_p}$    
    if inference~mode:
        $r^j$ = $A^F(r^j, \vartheta)$ 
    $s^j$, $u^j$ = (1 / $L_g$) * $\sum_{l=1}^{L_g} A_l^G(q^j, r^j, k^j)$    
    $\vartheta$ = (1 - 1/$j$) * $\vartheta$ + (1/$j$) * ($s^j$, $u^j$)       
\end{lstlisting}
\end{algorithm}

\paragraph{Proposition Phase.} 
A set of player (or proposer) agents ($\{A_i^P\}_{i=1}^{L_p}$) independently generate candidate responses to the same query. This design encourages diversity in reasoning paths and solution strategies, providing a rich pool of alternative perspectives for subsequent aggregation.

\paragraph{Inference Phase.} 
An aggregator (or fusion) agent($A^F$) synthesizes the candidate responses produced by the players. The aggregation is explicitly conditioned on each player’s historical performance profile, represented as a score with associated uncertainty. By incorporating this structured reliability information, the aggregator can weigh contributions beyond instantaneous agreement to inform consensus formation.

{\paragraph{Backward Phase.} A committee of LLM-based grader agents($\{A_l^G\}_{l=1}^{L_g}$) evaluates the players’ responses under task-specific or rubric-based criteria. These evaluations yield quantitative assessments of both quality and uncertainty, which are used to update the historical performance records of each player. Importantly, this phase does not modify the parameters of any underlying language model.}

\begin{figure*}[t]
    \centering
    \includegraphics[width=\textwidth]{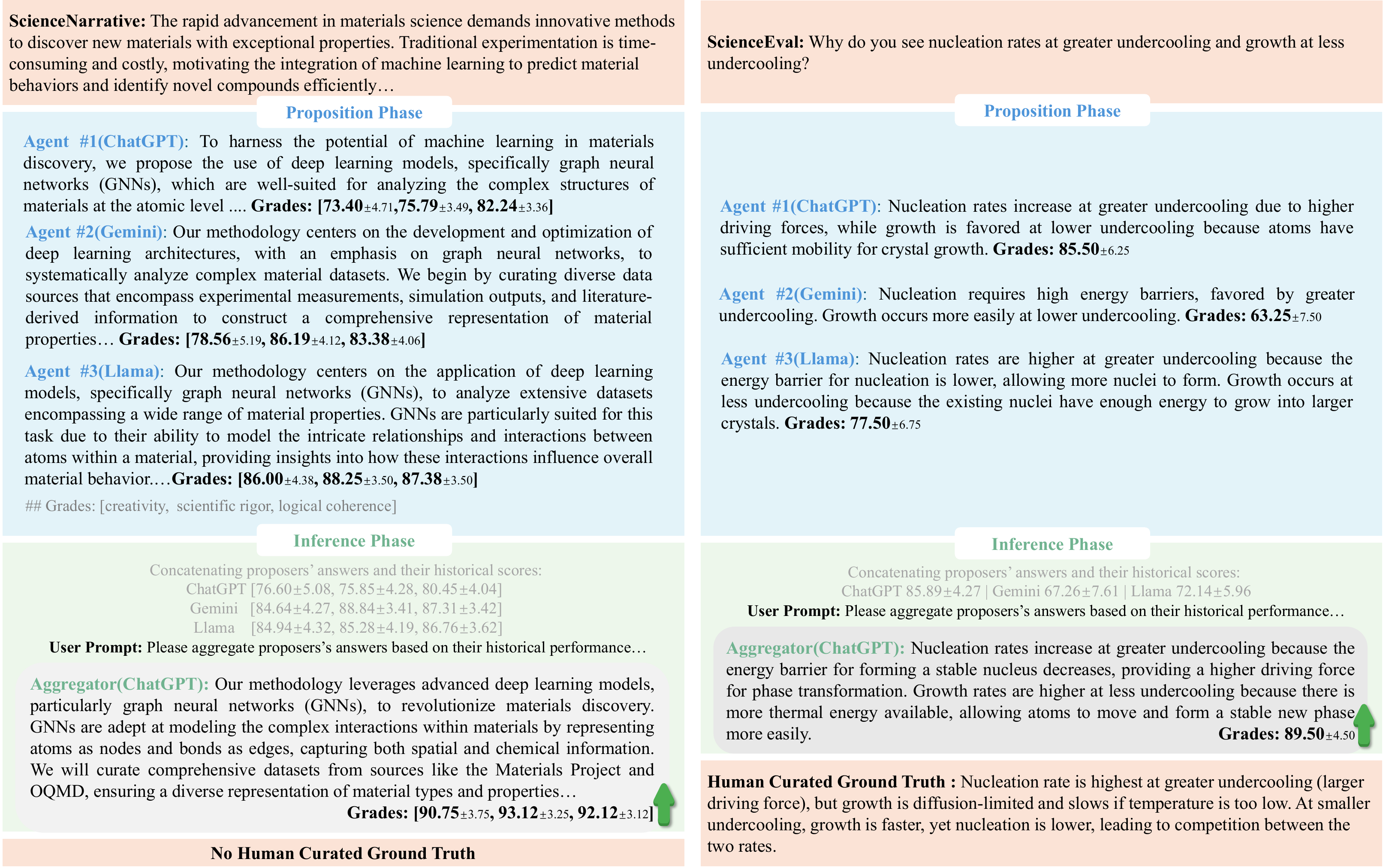}
    \caption{\label{fig:both_case} \footnotesize{\textbf{Example of a simplified \st\ and a subtask of \mt.}}
    }
\end{figure*}

\subsection{Encoding Historical Performance via Confidence-Weight Table}
In the backward phase, each evaluation yields a trust weight with two components: (i) a quality score in $[-100, 100]$, where negative values indicate incorrect or misleading responses, and (ii) an uncertainty value representing the confidence interval. Together, these form the equivalent of both a “grade” and a measure of “grading confidence,” ensuring that downstream decisions reflect not only evaluation outcomes but also their reliability. 

At round $j$, a query $q^j$ is uniformly sampled from the task dataset. The $i$-th player generate responses $r_i^j = A^P_i(q^j)$, which are then concatenated into $r^j$. Graders then evaluate these responses, producing $(s_l^j, u_l^j)$, which are averaged across $L_g$ graders to yield $(s^j, u^j)$. After $R$ rounds, the cumulative evaluations are distilled into a \textbf{confidence-weight table} $\vartheta = \Biggl(\sum_{j=1}^{R} s^j, \;\; \sum_{j=1}^{R} u^j\Biggr),$
capturing both the estimated quality and the associated uncertainty of each player. Intuitively, this table plays the role of a “track record” or “reputation system” for each player, ensuring that strong but inconsistent voices are weighted differently from consistently reliable ones. This design calibrates the committee’s trust in different players without retraining them, enabling adaptability and efficiency. 

The confidence-weight table $\vartheta \in \mathbb{R}^{L_p \times N}$ (index $L_p$ represents agent dimension, index $N$ represents  task dimension) serves as a structured memory of each player’s historical reliability across subtasks (\mt) or rubric dimensions (\st). Unlike fine-tuning approaches, \textsc{Roundtable Policy} does not alter the parameters of the underlying LLMs. Instead, $\vartheta$ is iteratively updated from repeated rounds of evaluation, much like how a committee gradually learns which members are most reliable in which topics.

\textbf{Inference.} Once $\vartheta$ has been learned, it guides consensus aggregation at the inference phase. Given a new query $q$, each player generates a response $r_i = A^P_i(q)$. The collection $r = \{r_1, \dots, r_{L_p}\}$ is passed to the fusion agent $A^F$, which aggregates them using the pre-constructed $\tilde{r} = A^F(r, \vartheta^{\text{pre-constructed}})$. The fusion process itself is not fine-tuned; it acts as an orchestrator that conditions on $\vartheta$ to produce a consensus answer. This mimics how a committee chair integrates both the content and credibility of reviewers before making a final decision.

\section{Results}
\label{sec:results}
\makeatletter
\begin{figure*}[t]
{\renewenvironment{table*}[1][]{\def\@captype{table}}{}%
 
\begin{table*}[t]
\centering

\caption{\footnotesize \textbf{Benchmark performance of single-agent baselines vs \textsc{Roundtable Policy} (RP).} }

\large
\renewcommand{\arraystretch}{0.8}
\resizebox{\textwidth}{!}{%
\begin{tabular}{ccccccccccc}
\toprule[1.5pt]
\raisebox{-1.5ex}{\large \textbf{Setting}} & \raisebox{-1.5ex}{\large \textbf{Model}} & \multicolumn{6}{c}{\large \textbf{Multi-Faceted Physical Reasoning}} & \multicolumn{3}{c}{\large \textbf{Other Domains}} \\
\cmidrule(lr){3-8} \cmidrule(lr){9-11}
 &  & \textbf{Mech. 1$\uparrow$} & \textbf{Mech. 2$\uparrow$} & \textbf{EE$\uparrow$} & \textbf{Opt.$\uparrow$} & \textbf{Thermo.$\uparrow$} & \textbf{Semi.$\uparrow$} & \textbf{Math$\uparrow$} & \textbf{Geo.$\uparrow$} & \textbf{Bio.$\uparrow$} \\
\midrule
\multirow{10}{*}{\rotatebox{90}{{\textbf{ScienceEval}}}} & GPT-4o \gpt & \cellcolor{lightblue}{\textbf{85.89\std{4.27}}} & \cellcolor{lightblue}{\textbf{78.26\std{5.37}}} & \cellcolor{lightblue}{\textbf{87.71\std{4.11}}} & 74.59\std{5.49} & \cellcolor{lightblue}{\textbf{85.14\std{4.40}}} & \cellcolor{lightblue}{\textbf{85.58\std{4.60}}} & 87.81\std{2.13} & \cellcolor{lightblue}{\textbf{83.55\std{2.20}}} & \cellcolor{lightblue}{\textbf{51.53\std{6.64}}} \\
 & Grok-2 \grok & 83.92\std{4.07} & 76.24\std{4.96} & 85.61\std{3.92} & \cellcolor{lightblue}{\textbf{76.18\std{4.99}}} & 82.02\std{4.10} & 82.63\std{4.39} & \cellcolor{lightblue}{\textbf{88.84\std{1.70}}} & 77.39\std{2.28} & 51.31\std{6.25} \\
 & Claude-3.5H \claude & 82.07\std{4.65} & 71.20\std{6.04} & 84.72\std{4.37} & 71.49\std{5.83} & 77.74\std{5.03} & 77.95\std{5.22} & 81.19\std{2.58} & 73.68\std{2.58} & 48.52\std{6.75} \\
 & GPT-4.5p \gpt & 72.86\std{6.77} & 71.46\std{6.96} & 76.67\std{6.12} & 72.78\std{6.36} & 77.18\std{5.71} & 77.68\std{6.17} & 82.50\std{2.68} & 82.59\std{2.08} & 50.63\std{6.89} \\
 & DeepSeek-chat \deepseek & 77.63\std{5.35} & 72.02\std{6.14} & 81.53\std{4.97} & 72.96\std{5.79} & 79.64\std{4.87} & 79.81\std{5.23} & 72.23\std{3.16} & 78.05\std{2.41} & 47.83\std{6.57} \\
 & Claude-3.7S \claude & 77.43\std{3.46} & 70.51\std{4.05} & 79.24\std{3.13} & 72.39\std{3.70} & 77.41\std{3.49} & 79.49\std{3.12} & 62.34\std{2.93} & 72.56\std{1.97} & 47.76\std{5.36} \\
 & Llama-3.3 \llama & 72.14\std{5.96} & 63.86\std{6.95} & 75.33\std{5.55} & 67.13\std{6.30} & 71.23\std{5.71} & 73.87\std{5.72} & 87.43\std{1.93} & 72.77\std{2.49} & 48.64\std{6.54} \\
 & Qwen-Plus \qwen & 80.32\std{5.01} & 72.36\std{6.03} & 82.97\std{4.71} & 73.42\std{5.66} & 81.49\std{4.70} & 81.18\std{4.93} & 86.99\std{2.08} & 76.27\std{2.48} & \cellcolor{superlightred}{\textbf{51.64\std{6.57}}} \\
 & Gemini-2.0 \gemini & 67.26\std{7.61} & 61.61\std{8.09} & 74.96\std{6.58} & 65.73\std{7.31} & 70.72\std{6.68} & 71.60\std{6.99} & 49.77\std{4.79} & 75.17\std{2.58} & 42.11\std{7.31} \\

\midrule
 & Ours\textsc{(RP)} & \cellcolor{superlightred}{\textbf{87.64\std{4.26}}} & \cellcolor{superlightred}{\textbf{81.50\std{5.36}}} & \cellcolor{superlightred}{\textbf{89.99\std{4.00}}} & \cellcolor{superlightred}{\textbf{81.29\std{4.96}}} & \cellcolor{superlightred}{\textbf{86.55\std{4.09}}} & \cellcolor{superlightred}{\textbf{91.82\std{4.34}}} & \cellcolor{superlightred}{\textbf{94.80\std{1.01}}} & \cellcolor{superlightred}{\textbf{85.33\std{1.63}}} & 47.55\std{7.02} \\
\midrule[1pt]
\raisebox{-1.5ex}{\large \textbf{Setting}} & \raisebox{-1.5ex}{\large \textbf{Model}} & \multicolumn{3}{c}{\large \textbf{Background}} & \multicolumn{3}{c}{\large \textbf{Methodology}} & \multicolumn{3}{c}{\large \textbf{Impact}} \\
\cmidrule(lr){3-5} \cmidrule(lr){6-8} \cmidrule(lr){9-11}
 &  & \textbf{Crea.$\uparrow$} & \textbf{Sci. Rig.$\uparrow$} & \textbf{Log. Coh.$\uparrow$} & \textbf{Crea.$\uparrow$} & \textbf{Sci. Rig.$\uparrow$} & \textbf{Log. Coh.$\uparrow$} & \textbf{Crea.$\uparrow$} & \textbf{Sci. Rig.$\uparrow$} & \textbf{Log. Coh.$\uparrow$} \\
\midrule
\multirow{10}{*}{\rotatebox{90}{ \textbf{ScienceNarrative} }} & GPT-4o \gpt & 70.99\std{5.01} & 74.73\std{4.68} & 82.75\std{3.76} & 75.49\std{5.23} & 74.81\std{5.03} & 79.63\std{4.31} & 72.15\std{5.19} & 70.53\std{5.25} & 79.64\std{4.19} \\
 & GPT-o3 \gpt & \cellcolor{lightblue}{\textbf{84.07\std{3.98}}} & 84.25\std{3.82} & 86.46\std{3.42} & \cellcolor{lightblue}{\textbf{85.92\std{3.95}}} & 86.70\std{3.56} & 85.84\std{3.56} & 83.90\std{4.15} & 81.52\std{4.25} & 84.86\std{3.69} \\
 & GPT-4.1 \gpt& 76.96\std{4.62} & 81.37\std{4.16} & 86.19\std{3.51} & 81.65\std{4.67} & 83.19\std{4.19} & 84.75\std{3.85} & 78.81\std{4.74} & 77.91\std{4.67} & 83.90\std{3.88} \\
 & GPT-4.1n \gpt & 77.39\std{4.55} & 80.94\std{4.15} & 85.59\std{3.53} & 80.48\std{4.69} & 82.30\std{4.19} & 83.97\std{3.85} & 77.97\std{4.74} & 77.22\std{4.65} & 82.94\std{3.90} \\
 & GPT-o3-m \gpt & 82.65\std{4.16} & \cellcolor{lightblue}{\textbf{84.78\std{3.83}}} & 87.26\std{3.36} & 83.15\std{4.35} & \cellcolor{lightblue}{\textbf{86.86\std{3.53}}} & 86.44\std{3.53} & 81.50\std{4.40} & 81.82\std{4.23} & 85.15\std{3.67} \\
 & GPT-4o-m \gpt & 72.78\std{4.94} & 74.64\std{4.70} & 82.19\std{3.80} & 75.44\std{5.16} & 76.18\std{4.85} & 79.81\std{4.23} & 72.78\std{5.18} & 70.69\std{5.24} & 78.82\std{4.25} \\
 & Grok-3 \grok & 78.50\std{5.10} & 79.10\std{4.50} & 85.50\std{3.64} & 80.96\std{4.75} & 81.26\std{4.47} & 83.58\std{3.99} & 78.81\std{4.78} & 76.34\std{4.93} & 83.04\std{4.01} \\
 & Claude-3.7S \claude & 82.47\std{4.22} & 84.73\std{3.92} & 88.17\std{3.28} & 85.52\std{4.20} & 86.65\std{3.79} & \cellcolor{lightblue}{\textbf{86.89\std{3.57}}} & \cellcolor{lightblue}{\textbf{84.53\std{4.22}}} & \cellcolor{lightblue}{\textbf{82.13\std{4.38}}} & 86.41\std{3.59} \\
 & DeepSeek-chat \deepseek & 84.05\std{4.20} & 83.74\std{4.12} & \cellcolor{lightblue}{\textbf{88.28\std{3.19}}} & 84.21\std{4.31} & 84.58\std{4.13} & 86.08\std{3.58} & 83.68\std{4.23} & 81.56\std{4.48} & \cellcolor{lightblue}{\textbf{86.50\std{3.46}}} \\
\midrule
 & Ours\textsc{(RP)} & \cellcolor{superlightred}{\textbf{90.25\std{3.95}}} & \cellcolor{superlightred}{\textbf{89.95\std{3.78}}} & \cellcolor{superlightred}{\textbf{91.04\std{3.09}}} & \cellcolor{superlightred}{\textbf{90.39\std{3.91}}} & \cellcolor{superlightred}{\textbf{91.28\std{3.35}}} & \cellcolor{superlightred}{\textbf{91.18\std{3.20}}} & \cellcolor{superlightred}{\textbf{90.15\std{3.95}}} & \cellcolor{superlightred}{\textbf{89.86\std{3.95}}} & \cellcolor{superlightred}{\textbf{90.99\std{3.26}}} \\
\bottomrule[1.5pt]
\end{tabular}
}
\label{tab:score_all}
\vspace{-5pt}
\end{table*}\par}
\vspace{6pt}
{\renewenvironment{figure*}[1][]{}{}%
 \input{figures/ablation/ablated_components}\par}
\end{figure*}
\makeatother

We evaluated \textsc{Roundtable Policy} (RP) in two complementary settings that reflect distinct but equally important demands of scientific work: answering pointwise questions across disciplines, and constructing structured, logically consistent arguments. In \mt, agents are simultaneously queried with nine domain-specific questions spanning geoscience~\citep{geobench}, biology~\citep{jin-etal-2019-pubmedqa}, mathematics~\citep{cobbe2021gsm8k}, and multi-faceted physical science, which contains six sub-categories (Supplementary Note~\sref{app:bench}). The tasks include multiple-choice, fill-in-the-blank, and short-answer formats, so the setting resembles a comprehensive written examination in which success requires broad and balanced reasoning rather than specialization in a single domain. In \st, agents generate structured scientific proposals following a standardized three-paragraph template consisting of \emph{Background}, \emph{Methodology}, and \emph{Impact} (Supplementary Note~\sref{app:bench}). This setting emphasizes the narrative dimension of scientific reasoning: beyond producing isolated facts, agents must organize ideas into a coherent, logically consistent and persuasive scientific narrative.

In each setting, a diverse set of state-of-the-art LLMs act as players and also serve as the single-agent baselines, following the notation in Supplementary Table~\sref{tab:model_notation}. We further compare RP with representative multi-agent frameworks:
\emph{Majority Vote}~\citep{wang2022self}, which aggregates agents by simple voting and serves as a canonical non-weighted baseline;
\emph{Weighted Vote}~\citep{taubenfeld-etal-2025-confidence}, which assigns confidence-based weights to individual agents;
\emph{Debate}~\citep{10.5555/3692070.3692537}, in which agents iteratively revise their answers through argument exchange;
\emph{Debate w Judge}~\citep{liang-etal-2024-encouraging}, which introduces an explicit adjudicator to select the final answer after debate;
\emph{Debate w Confidence}~\citep{chen-etal-2024-reconcile}, which further incorporates confidence signals to guide resolution;
and \emph{Mixture-of-Agents} (MoA)~\citep{wang2025mixtureofagents}, which constructs an agentic network to combine multiple agent outputs adaptively. Unless stated otherwise, gains are reported as the average relative improvement over the single-agent baselines.

\subsection{Structured consensus improves scientific reasoning across heterogeneous tasks}
\label{sec:performance}

We first asked whether the complementary capabilities of heterogeneous agents can be combined into answers that are better than those of the individual models. In \mt, RP obtained the highest score on eight of the nine tasks (Tasks~1--8; Table~\ref{tab:score_all}), which corresponds to a 13.01\% average improvement over the single-model baselines. The gains were especially pronounced on reasoning-heavy subtasks (multi-faceted physical reasoning). Because this average is taken over all nine baselines, it measures the improvement over a typical agent; relative to the strongest single model on each task, RP was higher by 1.41--6.24 points on Tasks~1--8. On some tasks the consensus was also more consistent: on Task~7 (Math), RP narrowed the confidence interval to $\pm 1.01$, compared with $\pm 1.70$--$4.79$ for the single-model baselines. Supplementary Figs.~\sref{fig:improvement_multitask} and~\sref{fig:improvement_singletask} visualize the fine-grained percentage improvements, and additional per-player and per-task analyses are provided in Supplementary Notes~\sref{sup:multi-performance} and~\sref{sup:single-performance}.

Biology (Task~9) was the exception. Here RP scored 47.55, below eight of the nine single models, whose scores all lie between 42.11 and 51.64. This is likely due to high intrinsic uncertainty and substantial disagreement among individual models, which limits the effectiveness of consensus-based aggregation. The contrast with Tasks~1--8 indicates that RP is most effective when model errors are diverse yet structured, and that the benefit of aggregation depends on the reliability of the underlying contributions.

The improvement extended from question answering to long-form scientific writing. In \st, RP achieved the highest score on all nine rubric dimensions, with an 11.04\% average improvement (Table~\ref{tab:score_all}). The gains concentrated in the \emph{Impact} section, particularly Scientific Rigor, while the overall scores remained tightly clustered around 90--91\%. Per-player analyses reveal right-skewed improvement distributions (Supplementary Note~\sref{sup:single-performance}): RP consistently lifts weaker proposals while preserving the quality of already strong drafts.

To test whether these gains hold at the level of individual evaluation rounds, we applied one-sided Wilcoxon signed-rank tests to the paired, round-level gains of RP over each single agent (Table~\ref{tab:significance}; Supplementary Note~\sref{app:significance}). RP significantly outperformed every agent in both settings at $\alpha = 0.05$, with $p$-values typically below $10^{-40}$. The per-round variance is substantial, so the advantage is not uniform across queries; over thousands of evaluations, however, the aggregated gains consistently converge to stable improvements. The advantage of RP is therefore not driven by isolated cases or short-lived fluctuations, but reflects a persistent distribution-level improvement. Fine-grained Wilcoxon results are reported in Supplementary Tables~\sref{tab:multitask_wilcoxon_1_5}--\sref{tab:singletask_wilcoxon_6_9}.

\begin{table}[H]
\centering

{\normalsize
\renewcommand{\arraystretch}{0.8}
\setlength{\tabcolsep}{3pt}

\caption{\footnotesize \textbf{Wilcoxon test results.} Rows show \textsc{Roundtable Policy}’s mean percentage gain over each baseline across all rubrics, with standard deviation and one-sided $p$-value. \checkmark marks significance at $\alpha = 0.05$.}
\label{tab:significance}

\begin{adjustbox}{max width=\linewidth, center}
\begin{tabular}{c cc cccccc}
\toprule
\textbf{Setting}  & \textbf{Model} & \textbf{Avg. Gain (\%)} & \textbf{Std (\%)} & \textbf{p-value$\downarrow$} & \textbf{Significance} \\
\midrule
\multirow{9}{*}{\rotatebox{90}{\textbf{ScienceEval}}} 
 &  \gpt-4o         & 6.476865  & 77.280369  & $0.0$                & \checkmark \\
 & \grok-2           & 5.254293  & 78.231667  & $1.0\times10^{-247}$ & \checkmark \\
 & \claude-3.5H    & 17.251754 & 101.839946 & $0.0$                & \checkmark \\
 & \gpt-4.5p   & 18.552776 & 94.463211  & $0.0$                & \checkmark \\
 & \deepseek-Chat     & 16.048551 & 96.935292  & $0.0$                & \checkmark \\
 & \claude-3.7S  & 13.732040 & 121.321577 & $4.4\times10^{-47}$  & \checkmark \\
 & \llama-3.3          & 24.988341 & 123.678543 & $0.0$                & \checkmark \\
 & \qwen-plus          & 12.586367 & 91.751670  & $0.0$                & \checkmark \\
 & \gemini-2.0        & 46.474125 & 145.071773 & $0.0$                & \checkmark \\
\midrule[1pt]
\multirow{9}{*}{\rotatebox{90}{\textbf{ScienceNarrative}}} 
 & \gpt-4o       & 27.087781 & 40.303119  & $0.0$ & \checkmark \\
 & \gpt-o3          & 13.316247 & 148.785675 & $0.0$ & \checkmark \\
 & \gpt-4.1           & 15.107889 & 73.600957  & $0.0$ & \checkmark \\
 & \gpt-4.1-nano     & 15.805949 & 28.188351  & $0.0$ & \checkmark \\
 & \gpt-o3-mini     & 9.687830  & 21.760829  & $0.0$ & \checkmark \\
 & \gpt-4o-mini     & 27.206693 & 42.707340  & $0.0$ & \checkmark \\
 & \grok-3           & 16.485277 & 29.024328  & $0.0$ & \checkmark \\
 & \claude-3.7S  & 7.761157  & 16.993190  & $0.0$ & \checkmark \\
 & \deepseek-Chat      & 7.165863  & 7.474061   & $0.0$ & \checkmark \\
\bottomrule
\end{tabular}
\end{adjustbox}
}
\end{table}  %

\subsection{Accumulated reliability records, more than aggregator strength, determine consensus quality}
\label{sec:stability}

What enables a heterogeneous group of agents to form a reliable consensus? Unlike voting or debate, RP conditions aggregation on a confidence-weight table that records the historical score of each agent together with its uncertainty. To examine which ingredients matter, we varied one component at a time: the number of player agents, the capability of the aggregator, the amount of historical evaluation distilled into the table, and the scoring strategy (Fig.~\ref{fig:ablated_components}).

Scaling the number of player agents led to steady performance improvements that gradually saturated, consistent with diminishing returns from additional candidate reasoning trajectories. In practice, the number of players is constrained by the maximum context window of the aggregator, as all proposals are concatenated for aggregation; we therefore evaluated scalability within this realistic operating regime. The capability of the aggregator mattered less. Ablating the aggregator from strong to weak resulted in bounded performance degradation rather than drastic collapse, indicating that RP does not rely on a single highly capable aggregator.

The amount of historical evaluation had the largest effect. Increasing the number of historical rounds used to distill the confidence-weight table (labelled `pretraining rounds' in Fig.~\ref{fig:ablated_components}) yielded consistent gains in both settings, which continued up to the largest number of rounds tested. With a single round, in contrast, the consensus fell below the average of the single-agent baselines. A poorly estimated table can thus mislead the aggregator, which suggests that the aggregator makes real use of the recorded reliability and that a more accurate encoding of historical reliability leads to better downstream performance. The table is constructed before inference and is not updated while queries are answered, so these gains reflect the quality of the stored record rather than adaptation at inference time.

Representing the uncertainty of the historical scores brought a further, smaller improvement: uncertainty-aware scoring outperformed vanilla score aggregation in both settings. These component-wise comparisons do not isolate a single causal factor, but together they suggest that effective collective reasoning depends not only on the diversity of the available proposals, but above all on how well the system has characterized their relative reliability.

\subsection{Reliability-aware consensus matches interaction-intensive collaboration at lower cost}
\label{sec:MAS}

\begin{table*}[h]
\centering
\caption{\footnotesize \textbf{Benchmark perfomance of multi-agent baselines vs \textsc{Roundtable Policy (RP)}.}}
\large
\renewcommand{\arraystretch}{1.0}
\resizebox{\textwidth}{!}{%
\begin{tabular}{ccccccccccc}
\toprule[1.5pt]
\raisebox{-1.5ex}{\large \textbf{Setting}} & \raisebox{-1.5ex}{\large \textbf{Method}} & \multicolumn{6}{c}{\large \textbf{Multi-Faceted Physical Reasoning}} & \multicolumn{3}{c}{\large \textbf{Other Domains}} \\
\cmidrule(lr){3-8} \cmidrule(lr){9-11}
 &  & \textbf{Mech. 1$\uparrow$} & \textbf{Mech. 2$\uparrow$} & \textbf{EE$\uparrow$} & \textbf{Opt.$\uparrow$} & \textbf{Thermo.$\uparrow$} & \textbf{Semi.$\uparrow$} & \textbf{Math$\uparrow$} & \textbf{Geo.$\uparrow$} & \textbf{Bio.$\uparrow$} \\
\midrule
\multirow{6}{*}{\rotatebox{90}{{ \textbf{ScienceEval} }}} & Majority Vote~\citep{wang2022self} 
& 82.48\std{4.51} & 74.02\std{5.63} & 85.01\std{4.25} & 73.65\std{5.48} & 81.01\std{4.56} & 81.68\std{4.77} & 85.93\std{2.12} & 79.03\std{2.27} & 50.88\std{6.54} \\

 & Weighted Vote ~\citep{taubenfeld-etal-2025-confidence} 
& 81.56\std{4.58} & 74.46\std{5.54} & 84.24\std{4.29} & 73.13\std{5.45} & 81.76\std{4.49} & 81.16\std{4.77} & 83.03\std{2.27} & 76.56\std{2.31} & 49.28\std{6.48} \\

 & Debate ~\citep{10.5555/3692070.3692537} 
 & 85.68\std{4.12} & 80.34\std{4.28} & 88.01\std{4.02} & 77.25\std{3.90} & 84.97\std{3.30} & 89.17\std{4.09} & 92.23\std{1.05} & 84.05\std{1.90} & \cellcolor{lightblue}{\textbf{51.68\std{6.25}}} \\

 & Debate w Judge ~\citep{liang-etal-2024-encouraging} 
 & 86.93\std{4.28} & 80.36\std{4.20} & 88.42\std{4.10} & 76.25\std{4.05} & 85.26\std{3.77} & 89.43\std{4.25} & 92.04\std{1.25} & 83.65\std{1.97} & 51.23\std{6.07} \\

 & Debate w Confidence ~\citep{chen-etal-2024-reconcile} 
 & 86.24\std{3.50} & 79.61\std{3.69} & 88.15\std{3.55} & 79.39\std{3.73} & 86.27\std{3.02} & 89.95\std{3.45} & \cellcolor{lightblue}{\textbf{93.18\std{1.55}}} & 83.92\std{1.30} & 49.17\std{4.75} \\

 & Mixture-of-Agents ~\citep{wang2025mixtureofagents} 
 & \cellcolor{lightblue}{\textbf{87.53\std{3.75}}}  & \cellcolor{lightblue}{\textbf{81.02\std{3.74}}}  & \cellcolor{lightblue}{\textbf{89.74\std{3.47}}}  & \cellcolor{lightblue}{\textbf{80.68\std{3.23}}}  & \cellcolor{superlightred}{\textbf{87.29\std{3.05}}}  & \cellcolor{lightblue}{\textbf{91.25\std{3.59}}}  & 93.05\std{1.42}  & \cellcolor{lightblue}{\textbf{85.28\std{1.25}}}  & \cellcolor{superlightred}{\textbf{53.50\std{5.31}}} \\
 
\midrule

 & \textsc{Roundtable Policy} & \cellcolor{superlightred}{\textbf{87.64\std{4.26}}} & \cellcolor{superlightred}{\textbf{81.50\std{5.36}}} & \cellcolor{superlightred}{\textbf{89.99\std{4.00}}} & \cellcolor{superlightred}{\textbf{81.29\std{4.96}}} & \cellcolor{lightblue}{\textbf{86.55\std{4.09}}} & \cellcolor{superlightred}{\textbf{91.82\std{4.34}}} & \cellcolor{superlightred}{\textbf{94.80\std{1.01}}} & \cellcolor{superlightred}{\textbf{85.33\std{1.63}}} & 47.55\std{7.02} \\

\midrule[1pt]
\raisebox{-1.5ex}{\large \textbf{Setting}} & \raisebox{-1.5ex}{\large \textbf{Model}} & \multicolumn{3}{c}{\large \textbf{Background}} & \multicolumn{3}{c}{\large \textbf{Methodology}} & \multicolumn{3}{c}{\large \textbf{Impact}} \\
\cmidrule(lr){3-5} \cmidrule(lr){6-8} \cmidrule(lr){9-11}
 &  & \textbf{Crea.$\uparrow$} & \textbf{Sci. Rig.$\uparrow$} & \textbf{Log. Coh.$\uparrow$} & \textbf{Crea.$\uparrow$} & \textbf{Sci. Rig.$\uparrow$} & \textbf{Log. Coh.$\uparrow$} & \textbf{Crea.$\uparrow$} & \textbf{Sci. Rig.$\uparrow$} & \textbf{Log. Coh.$\uparrow$} \\
\midrule
\multirow{6}{*}{\rotatebox{90}{\small \textbf{ScienceNarrative} }} & Majority Vote ~\citep{wang2022self} 
& 81.37\std{4.30} & 82.99\std{4.07} & 86.96\std{3.41} & 83.46\std{4.37} & 84.64\std{3.98} & 85.43\std{3.70} & 81.93\std{4.41} & 80.11\std{4.51} & 84.90\std{3.73} \\

 & Weighted Vote ~\citep{taubenfeld-etal-2025-confidence} 
& 81.47\std{4.30} & 83.22\std{4.05} & 87.26\std{3.36} & 83.74\std{4.36} & 84.90\std{3.97} & 85.74\std{3.68} & 82.37\std{4.38} & 80.35\std{4.51} & 85.22\std{3.70} \\

 & Debate ~\citep{10.5555/3692070.3692537} 
 & 83.16\std{4.88} & 84.99\std{5.25} & 88.25\std{3.64} & 86.14\std{5.25} & 87.20\std{3.50} & 86.92\std{3.25} & 85.11\std{4.02} & 83.65\std{4.00} & 85.96\std{3.75} \\

 & Debate w Judge ~\citep{liang-etal-2024-encouraging} 
 & 86.75\std{3.75} & 85.50\std{4.25} & 86.04\std{4.17} & 87.05\std{3.50} & 88.96\std{3.25} & 88.81\std{3.10} & 87.25\std{4.06} & 83.50\std{3.85} & 85.41\std{3.71} \\

 & Debate w Confidence ~\citep{chen-etal-2024-reconcile} 
 & 86.47\std{3.20} & \cellcolor{lightblue}{\textbf{89.25\std{3.34}}} & \cellcolor{lightblue}{\textbf{90.02\std{2.95}}} & 86.99\std{3.13} & \cellcolor{lightblue}{\textbf{90.25\std{3.25}}} & \cellcolor{lightblue}{\textbf{90.70\std{2.75}}} & 87.08\std{3.50} & \cellcolor{lightblue}{\textbf{88.63\std{3.31}}} & \cellcolor{lightblue}{\textbf{89.82\std{2.95}}} \\

 & Mixture-of-Agents ~\citep{wang2025mixtureofagents} 
 & \cellcolor{lightblue}{\textbf{89.98\std{3.90}}} & 86.17\std{4.47} & 89.50\std{3.68} & \cellcolor{lightblue}{\textbf{90.25\std{3.65}}} & 88.92\std{3.25} & 90.15\std{3.57} & \cellcolor{lightblue}{\textbf{89.75\std{4.25}}} & 88.60\std{3.93} & 89.22\std{3.10} \\

\midrule

 & \textsc{Roundtable Policy} & \cellcolor{superlightred}{\textbf{90.25\std{3.95}}} & \cellcolor{superlightred}{\textbf{89.95\std{3.78}}} & \cellcolor{superlightred}{\textbf{91.04\std{3.09}}} & \cellcolor{superlightred}{\textbf{90.39\std{3.91}}} & \cellcolor{superlightred}{\textbf{91.28\std{3.35}}} & \cellcolor{superlightred}{\textbf{91.18\std{3.20}}} & \cellcolor{superlightred}{\textbf{90.15\std{3.95}}} & \cellcolor{superlightred}{\textbf{89.86\std{3.95}}} & \cellcolor{superlightred}{\textbf{90.99\std{3.26}}} \\
 
\bottomrule[1.5pt]
\end{tabular}
}
\label{tab:other_mas}
\end{table*}
\begin{table}[h]
\centering
\caption{\textbf{Token cost per task vs average improvement.}}
\label{tab:e2_efficiency}
\small
\begin{tabular}{lcc|cc}
\toprule
& \multicolumn{2}{c|}{\textbf{\mt}} & \multicolumn{2}{c}{\textbf{\st}} \\
\cmidrule(lr){2-3} \cmidrule(lr){4-5}
\textbf{Method} & Token & Gain & Token & Gain \\
\midrule
\rowcolor{gray!8}
{Ours (RP)} & 5.36k & \cellcolor{lightblue}{+13.01\%} & 6.95k & \cellcolor{superlightred}{+11.04\%} \\
\midrule
Majority & 4.83k & +5.02\% & 6.26k & +2.42\% \\
Weighted & 4.83k & +3.73\% & 6.26k & +2.75\% \\
\midrule
Debate & 14.51k & +11.03\% & 18.88k & +5.09\% \\
Debate w Conf. & 14.62k & +11.41\% & 18.97k & +8.88\% \\
Debate w Judge & 15.02k & +11.06\% & 19.46k & +6.16\% \\
\midrule
MoA & 15.31k & \cellcolor{superlightred}{+13.44\%} & 19.75k & \cellcolor{lightblue}{+9.33\%} \\
\bottomrule
\end{tabular}
\end{table}
If reliability information carries much of the benefit, is elaborate interaction between agents necessary? We compared RP with representative multi-agent frameworks (implementation details in Supplementary Note~\sref{app:mas}), including voting-based aggregation~\citep{wang2022self,taubenfeld-etal-2025-confidence}, debate-style interaction~\citep{10.5555/3692070.3692537, liang-etal-2024-encouraging, chen-etal-2024-reconcile}, and agentic-network-style aggregation~\citep{wang2025mixtureofagents}. These frameworks differ in how agent influence is assigned, updated, and stabilized during aggregation.

In \mt, RP attained the best or second-best result on 8 of the 9 task dimensions (best on seven; Table~\ref{tab:other_mas}), with particularly strong performance on reasoning-intensive domains such as multi-faceted physical reasoning and Math. Mixture-of-Agents was the closest alternative: it scored highest on Thermodynamics and Biology, and its average improvement (+13.44\%) slightly exceeded that of RP (+13.01\%; Table~\ref{tab:e2_efficiency}). Interaction-heavy baselines such as Debate and Debate w Confidence were competitive, and on individual tasks even superior. The differences among RP, Mixture-of-Agents and the debate variants are small relative to the reported confidence intervals, so we regard these methods as comparable in this setting. Voting-based methods, whose aggregation is uniform and memoryless, improved considerably less (+5.02\% and +3.73\%). Biology remained the weak point of RP, which scored lowest among all multi-agent methods on this task.

\begin{figure*}[t]
\centering
\includegraphics[width=0.95\textwidth]{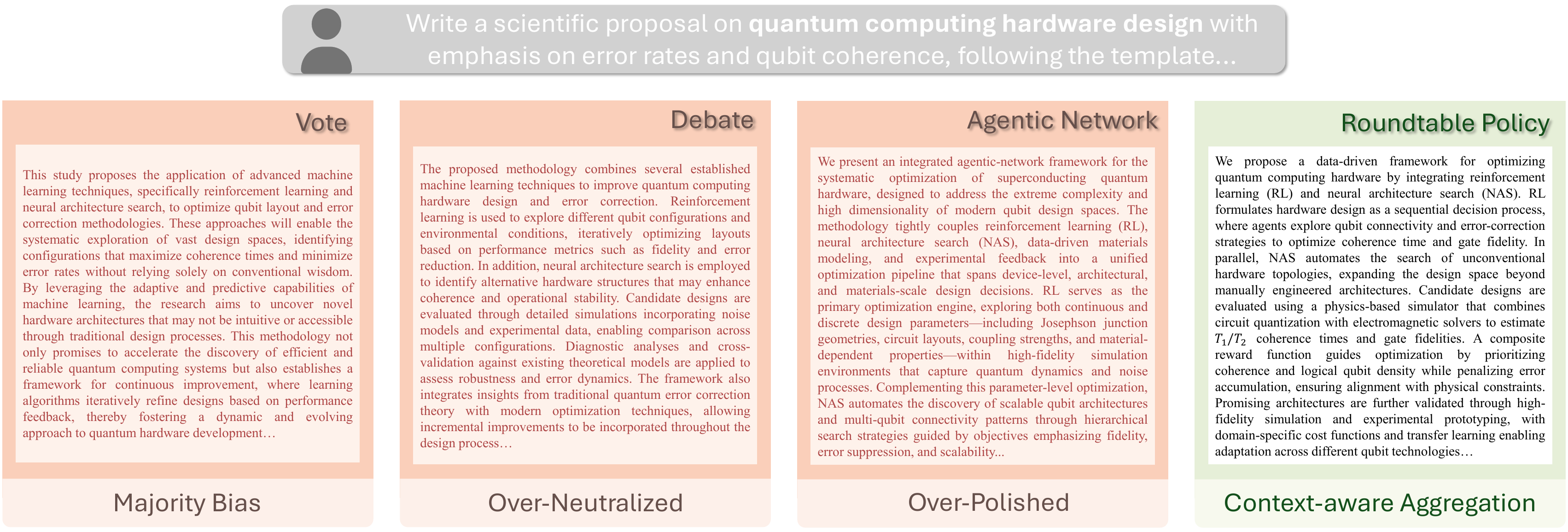}
\caption{\label{fig:mas_example}
\footnotesize{\textbf{Qualitative examples of multi-agent systems vs \textsc{Roundtable Policy}.}}}
\end{figure*}

In \st, RP achieved the best performance across all nine rubrics (Table~\ref{tab:other_mas}), although its margins over the second-best method were small (0.14--1.23 points). The competing methods exhibited strengths in specific categories: Debate w Confidence ranked second on every scientific-rigor and logical-coherence dimension, whereas Mixture-of-Agents ranked second on all three creativity dimensions. Debate-based methods thus do well in dimensions emphasizing argumentative clarity or coherence, but their performance degrades when tasks require creativity; incorporating confidence cues during interaction mitigates this only partially. RP, by comparison, maintained balanced performance across all rubrics. This balance is consistent with its structured, long-term record of each agent's strengths, although our experiments do not isolate this factor. Fig.~\ref{fig:mas_example} illustrates the qualitative differences on an example proposal.

The clearest difference lies in cost. As shown in Table~\ref{tab:e2_efficiency}, RP delivered competitive or superior performance gains at substantially lower token cost than debate-based and agentic-network-style methods. In \mt, RP achieved a +13.01\% improvement with only 5.36k tokens per task, whereas debate-style methods and Mixture-of-Agents consumed nearly three times more tokens (14.51k--15.31k) for comparable gains. Similarly, in \st, RP attained the largest improvement (+11.04\%) with 6.95k tokens, against 18.88k--19.75k for Debate and MoA. Voting was slightly cheaper (4.83k and 6.26k tokens) but delivered less than half of the improvement. Structured reliability information can therefore replace much of the interaction on which these protocols rely, yielding a favorable performance--cost trade-off.

\subsection{The advantage persists under independent graders}
\label{sec:graders}

\begin{table*}[t]
\centering
\caption{\textbf{Re-grading performance with diverse graders.}}
\label{tab:re-grading}
\small
\begin{adjustbox}{width=\linewidth}
\begin{tabular}{lcccc|cccc}
\toprule
& \multicolumn{4}{c|}{\textbf{\mt}} & \multicolumn{4}{c}{\textbf{\st}} \\
\cmidrule(lr){2-5} \cmidrule(lr){6-9}
\textbf{Grader Base Model} & RP & MoA & Majority & Debate & RP & MoA & Majority & Debate \\
\midrule
SmolLM3-3B & \cellcolor{superlightred}{+10.62\%} & \cellcolor{lightblue}{+10.58\%} & +3.15\% & +9.15\% & \cellcolor{superlightred}{+7.31\%} & +4.28\% & +2.12\% & \cellcolor{lightblue}{+6.73\%} \\
Phi-4-mini-reasoning & \cellcolor{superlightred}{+15.83\%} & \cellcolor{lightblue}{+12.75\%} & +5.97\% & +10.99\% & \cellcolor{superlightred}{+8.56\%} & \cellcolor{lightblue}{+7.85\%} & +3.19\% & +7.47\% \\
Gemma-3-4B & \cellcolor{superlightred}{+6.63\%} & +5.86\% & +1.82\% & \cellcolor{lightblue}{+6.49\%} & \cellcolor{lightblue}{+3.25\%} & \cellcolor{superlightred}{+4.46\%} & +1.34\% & +2.75\% \\
OLMo-2-0425-1B-Instruct & \cellcolor{superlightred}{+9.45\%} & \cellcolor{lightblue}{+9.40\%} & +3.71\% & +8.95\% & \cellcolor{lightblue}{+4.72\%} & \cellcolor{superlightred}{+5.96\%} & +1.84\% & +3.96\% \\
\bottomrule
\end{tabular}
\end{adjustbox}
\end{table*}

\begin{figure*}[t]
\centering
\includegraphics[width=\linewidth]{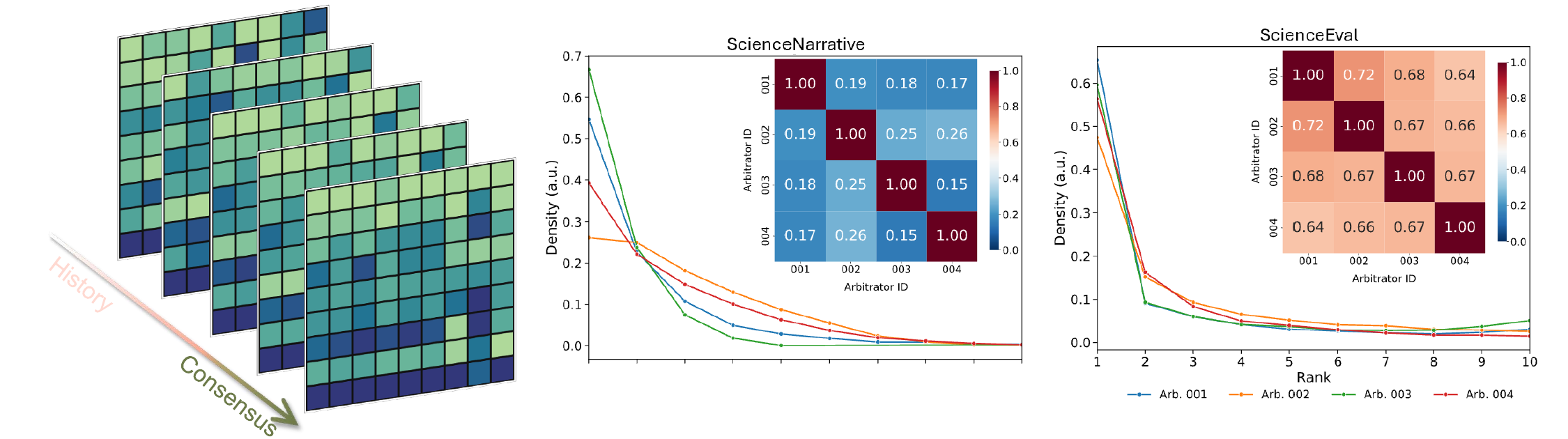}
\vspace{-5pt}
\caption{\label{fig:bias}
\footnotesize{
\textbf{Consensus of the grader ensemble.}\emph{Left:} Committee's consensus visualized as the convergence of the confidence-weight matrix.
\emph{Right:} Consistency of AI graders.
The outer plots represent the rank distributions given by different arbitrators; the inset heatmaps report pairwise inter-grader agreement by Kendall’s Tau.
}
}
\end{figure*}

All of the scores above were assigned by a committee of LLM graders, and the same kind of committee evaluation produces the historical record from which the confidence-weight table is built. The comparison could therefore favor RP if its consensus merely matched the preferences of one particular set of graders. We examined this possibility in two ways.

First, we re-evaluated the outputs with four additional grader base models ranging from 1B to 4B scale (Table~\ref{tab:re-grading}). RP consistently achieved the largest or second-largest relative improvement among the multi-agent methods across both \mt~and \st: it ranked first under all four graders in \mt, and first under two graders and second to MoA under the other two in \st. The magnitude of the improvements differed considerably between graders, but the ranking remained stable even under these weaker graders, indicating that the advantage of RP is not an artifact of a particular evaluator.

Second, we analyzed the consistency of the grader committee itself, using (i) the rank distributions assigned to RP and (ii) pairwise agreement measured by Kendall’s Tau (Supplementary Note~\sref{app:kendall}). Aggregate results are shown in Fig.~\ref{fig:bias}, with detailed breakdowns in Supplementary Note~\sref{sup:graders}. In \mt, graders exhibit concentrated rank distributions and high Kendall’s Tau values (0.64--0.72), indicating strong agreement on objective tasks. In \st, inter-grader agreement decreases markedly (0.15--0.26), particularly for subjective dimensions such as creativity and logical coherence. Agreement among graders does not by itself establish that their judgments are correct, and the weaker agreement on open-ended writing means that absolute scores in \st~should be interpreted with more caution than those in \mt. It also argues for scoring with a committee of graders rather than relying on a single one.

\section{Related Work}
\textbf{Multi-Agent Systems.} Early approaches show that majority~\citep{wang2022self} and confidence-weighted schemes~\citep{taubenfeld-etal-2025-confidence} can improve reliability by combining diverse reasoning trajectories. Besides, Du et al.~\citep{10.5555/3692070.3692537} present multi-agent debate, while Liang et al.~\citep{liang-etal-2024-encouraging} introduce asymmetric debater–judge roles in debate. Further collaborative paradigm emphasize fine-grained interaction protocols~\citep{chan2024chateval, zhang-etal-2024-exploring,wang2025mixtureofagents}, although debate does not reliably outperform other prompting strategies~\citep{smit2024mad}. Stream of Belief~\citep{yao2026streamofbelief} moves the exchange into latent space, where agents refine continuous beliefs until the collective state converges.
\textsc{Roundtable Policy} instead conditions a single black-box fusion step on an explicit record of each agent's measured reliability. Our work thus provides a complementary perspective by studying how reliability-aware consensus can be used during inference, compatible with other collaboration strategies.

\textbf{Reliability Estimation and LLM Judges.}
Weighting fallible sources by their estimated reliability dates back to Dawid and Skene~\citep{dawid1979maximum}. For LLMs, verbalized confidence tends to be overconfident~\citep{xiong2024uncertainty}, which motivates our use of externally graded history. LLM judges are scalable but favor their own generations~\citep{panickssery2024selfpreference}, one reason we score with a committee of graders rather than a single judge. VerifiedLab~\citep{yao2026verifiedlab} similarly checks each claim against the measured reliability of its underlying records.

\textbf{LLMs for Science.}
Large foundation models have driven broad scientific advances, from protein structure prediction~\citep{senior2020improved, alphafold, doi:10.1126/science.abj8754, doi:10.1126/science.ade2574} to domain-specialized tasks spanning math, chemistry, biology, and geoscience~\citep{romera2024mathematical, alphageometry2024, yue2024mammoth, yu2024llasmol, labrak-etal-2024-biomistral, li-etal-2023-geolm}. LLM-based multi-agent pipelines also curate scientific knowledge at scale, for instance by text-mining device records from the perovskite solar cell literature~\citep{dong2026perovskitenet}. Our work complements increasingly complex agentic workflows that couple LLMs with external tools and experimental pipelines~\citep{yao2023react, boiko2023autonomous, lu2024aiscientistfullyautomated, yamada2025} by focusing on end-to-end inference-time reasoning and aggregation.

\textbf{Benchmarks for Scientific Reasoning.} 
Scientific reasoning has been widely benchmarked. Math GSM8K~\citep{cobbe2021gsm8k} focus on step-by-step numerical reasoning, biomedical PubMedQA~\citep{jin-etal-2019-pubmedqa} assess Q\&A grounded in literature, GPQA~\citep{rein2024gpqa} and AlphaGeometry~\citep{alphageometry2024} introduce problem-solving in specialized settings. Multi-domain benchmarks such as MMLU~\citep{hendryckstest2021} and BIG-Bench~\citep{srivastava2023beyond} evaluate general reasoning across a broad range of technical subjects, ARC~\citep{clark2018think}, GSM-Symbolic~\citep{mirzadeh2025gsmsymbolic}, and GPQA-Diamond~\citep{rein2024gpqa} further emphasize compositional reasoning in diverse settings. Morph-Bench~\citep{yao2026morphbench} probes whether vision--language models can ground microscopy images in physical scale. Our work extends these efforts by integrating heterogeneous, long-context scientific reasoning, providing explicit dimensions for assessing expertise in multi-agent systems.

\section{Discussion}

We presented \textsc{Roundtable Policy}, an inference-time reasoning framework for multi-agent systems, and showed that it improves performance over single agents across two scientific benchmarks while matching interaction-intensive protocols at roughly a third of their token cost. By decoupling influence assignment from transient interaction dynamics and grounding it in long-term, fine-grained measurement, RP enables stable context-aware aggregation. Beyond increasingly complex interaction protocols, multi-agent systems also benefit from mechanisms that explicitly model and stabilize agent reliability across downstream settings, motivating more principled aggregation strategies and task-aware evaluation frameworks. Future inference-time reasoning frameworks for multi-agent systems may increasingly view collaboration as a problem of reliability modeling and consensus formation, rather than relying solely on more elaborate interaction dynamics.

\textbf{Limitations.} \textsc{Roundtable Policy} relies on LLM-based graders, which may introduce bias or calibration errors: such judges are known to favor their own generations~\citep{panickssery2024selfpreference}, and agreement among our graders was weak on open-ended writing. A grader committee and re-grading with independent models reduce this concern but do not remove it. Its confidence-weight table assumes relatively stable agent reliability across tasks, which may not always hold in dynamic domains. Aggregation also cannot correct errors that the agents share, as the Biology task shows. RP is also more costly than single-agent inference, and while it improves aggregation-level interpretability, it does not fully explain the internal reasoning behind generated outputs. Finally, both benchmarks are text-only, whereas scientific evidence is often visual and tied to physical scale~\citep{yao2026morphbench}.

\textbf{Future Work.}
Several directions can further extend the \textsc{Roundtable Policy} framework.
First, future work may explore adaptive and task-conditioned reliability modeling, allowing the confidence-weight table to dynamically adjust to domain shifts or evolving agent behaviors.
Second, integrating external verification signals, such as records verified field by field~\citep{yao2026verifiedlab}, or human-in-the-loop feedback could improve calibration and mitigate potential grader bias.
Third, more efficient aggregation mechanisms—such as compressed proposal encoding, hierarchical summarization, or routing queries to a subset of agents~\citep{chen2024frugalgpt}—may further reduce token cost and improve scalability under strict context window constraints.
Finally, extending RP beyond static aggregation toward iterative or interactive multi-round settings, including exchange in latent space~\citep{yao2026streamofbelief}, may enable richer forms of collaborative reasoning while preserving stability.

\textbf{Broader Impact.} \textsc{Roundtable Policy} can improve multi-agent performance across a wide range of downstream tasks, particularly in complex scientific reasoning and structured knowledge synthesis, by providing a stable and efficient mechanism for aggregating diverse agent perspectives.

\section*{Acknowledgement}

JL acknowledges support by the Defense Advanced Research Projects Agency (DARPA) under Agreement No. HR00112490369. The content is solely the responsibility of the authors and does not necessarily represent the official views of the funding
entities.

\bibliographystyle{naturemag-linked}
{\footnotesize\hypersetup{urlcolor=black}\bibliography{references}}

\clearpage
\onecolumn
\setcounter{secnumdepth}{1}
\renewcommand{\thesection}{\Alph{section}}
\setcounter{section}{0}
\titleformat{\section}{\sffamily\bfseries\fontsize{11}{13}\selectfont}{Appendix \thesection\ \textbar}{0.4em}{}
\crefname{section}{Appendix}{Appendices}\Crefname{section}{Appendix}{Appendices}

\section{Detailed Demonstration of \textsc{Roundtable Policy}}
\label{app:detailed_demo}

\begin{figure}
    \centering
    \includegraphics[width=\textwidth,height=0.93\textheight,keepaspectratio]{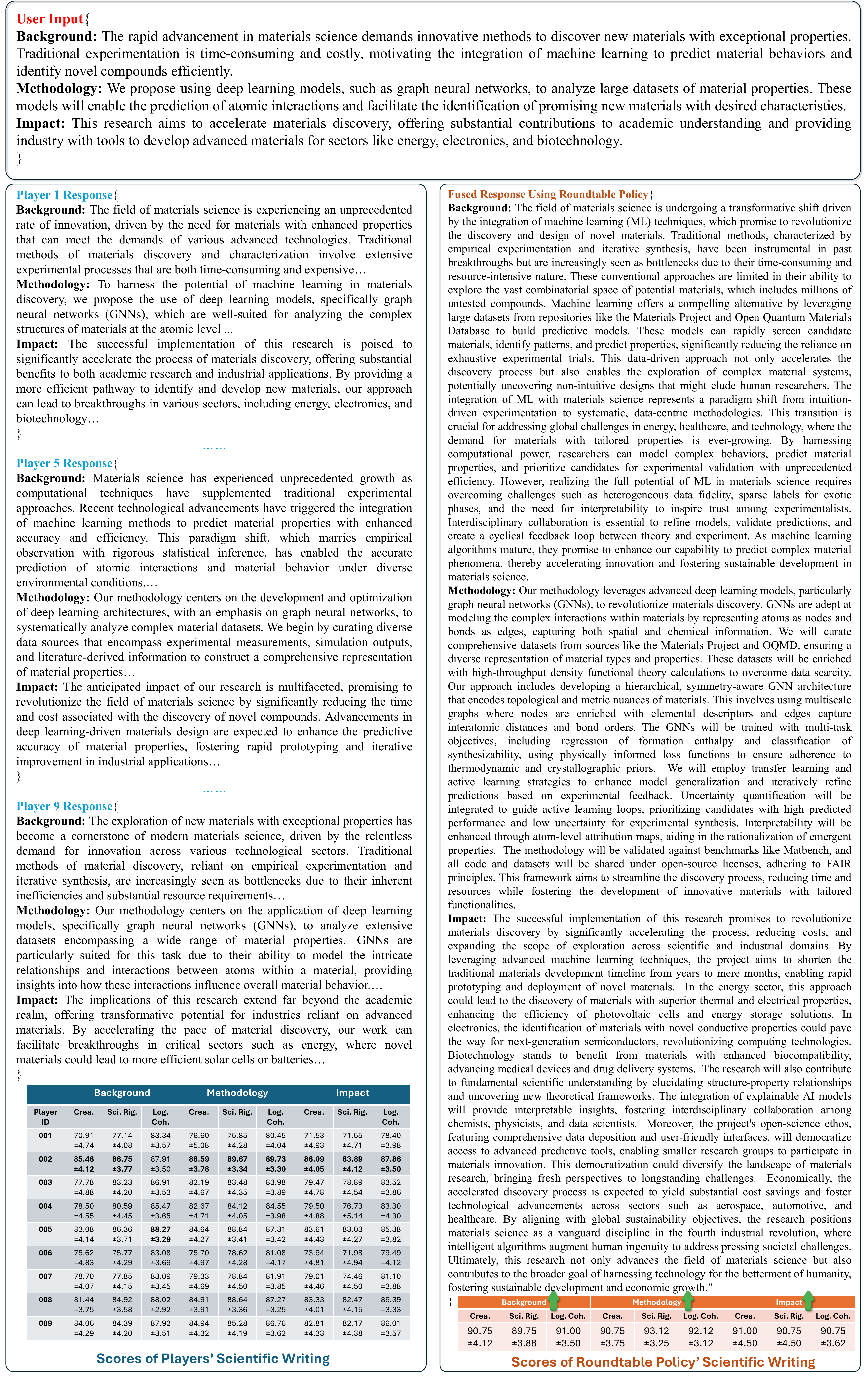}
    \caption{\label{fig:singletask_demo} \textbf{Demonstration of \st.}}
\end{figure}

\begin{figure}
    \centering
    \includegraphics[width=\textwidth,height=0.93\textheight,keepaspectratio]{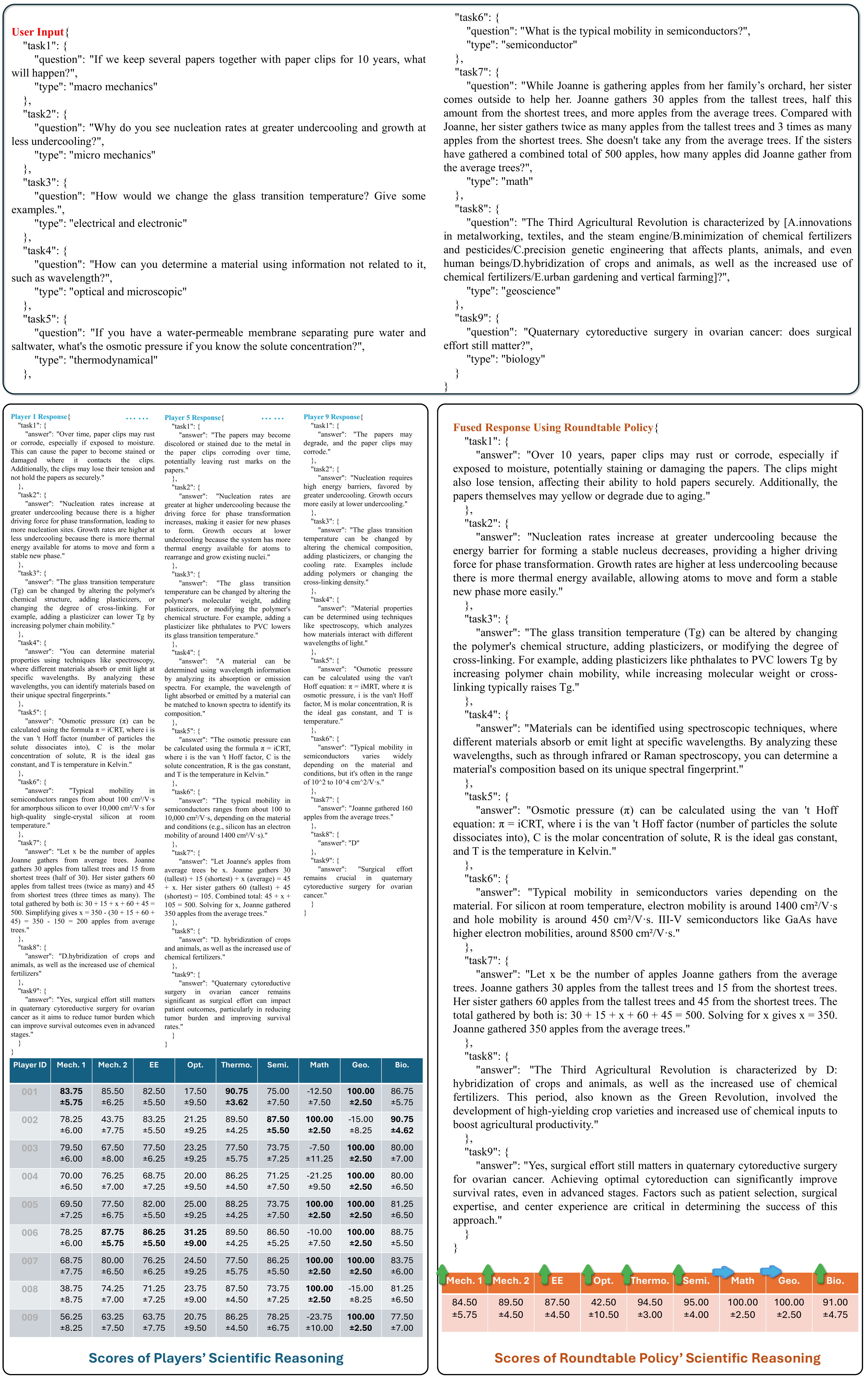}
    \caption{\label{fig:multitask_demo} \textbf{Demonstration of \mt.}}
\end{figure}

To provide a concrete illustration of how \textsc{Roundtable Policy} (RP) operates in practice, we present two representative demonstrations: one in the \mt~regime (Fig.~\ref{fig:multitask_demo}) and one in the \st~regime (Fig.~\ref{fig:singletask_demo}). These examples showcase how RP transforms fragmented outputs from multiple LLMs into coherent, factually grounded, and narratively rigorous responses.

\paragraph{\mt~ demonstration.}  
Fig.~\ref{fig:multitask_demo} presents a representative \mt~ round, where nine LLM \emph{players} respond simultaneously to a heterogeneous set of scientific queries spanning \emph{mechanics, thermodynamics, semiconductor physics, mathematics, geoscience, and biology}. Individual player outputs illustrate the limitations of single-model reasoning: some provide only partial explanations (e.g., noting that paper clips corrode without addressing stress relaxation), others introduce inaccuracies (e.g., oversimplifying nucleation–growth dynamics), and several fail to maintain consistency across domains (e.g., contradictory accounts of polymer glass transition). These fragmented outputs reflect the heterogeneity of LLM competencies—stronger players excel in certain areas, while weaker ones often produce vague or under-specified responses.  

By contrast, the fused response generated by RP demonstrates marked improvements in both \emph{factual coverage} and \emph{reasoning depth}. For instance, RP not only describes corrosion of paper clips but also contextualizes long-term degradation and mechanical implications; in the nucleation–growth problem, RP integrates complementary insights into thermodynamic driving forces and kinetic limitations, producing a textbook-quality explanation. Similarly, in the polymer glass transition question, RP combines molecular structure, chain length, and plasticizer effects into a unified account that no single player captured in full.  

Committee grading further substantiates these gains: RP exhibits higher factual accuracy and reduced uncertainty intervals across tasks. Improvements are particularly pronounced in symbolic and reasoning-intensive domains such as thermodynamics and mathematics, where RP delivers correct solutions while clarifying intermediate reasoning steps. These results highlight how RP balances correctness and clarity across heterogeneous domains.

\paragraph{\st~ demonstration.}  
Fig.~\ref{fig:singletask_demo} illustrates the \st~ regime, where each player generates a three-paragraph scientific proposal (\emph{Background, Methodology, Impact}). Individual drafts vary widely in creativity, methodological depth, and logical flow. Some emphasize novelty but lack technical detail; others offer rigorous methods but fail to contextualize the problem or articulate downstream impact. Such stylistic and structural inconsistencies exemplify the difficulty of producing a coherent \emph{scientific narrative} with a single LLM.  

In contrast, the fused proposal produced by RP exhibits a markedly stronger narrative arc. The \emph{Background} situates the research problem within its broader scientific context; the \emph{Methodology} clearly articulates technical contributions with precision; and the \emph{Impact} section highlights both scientific significance and societal relevance. Grading results confirm consistent improvements across all rubric dimensions—creativity, scientific rigor, and logical coherence—demonstrating that RP effectively mitigates individual stylistic biases while amplifying collective strengths. Even strong baseline players benefit: while their drafts are often polished in one dimension, RP integrates complementary insights to deliver uniformly high-quality writing.  

\paragraph{Summary.}  
Together, these demonstrations show how \textsc{Roundtable Policy} transforms fragmented or uneven model outputs into coherent, high-quality responses. In the \mt~ regime, RP enhances cross-domain reasoning reliability by integrating complementary knowledge across scientific fields. In the \st~ regime, RP substantially improves narrative structure and explanatory depth in long-form proposal writing. These examples reinforce the central claim that RP enables LLM ensembles to achieve more factual, rigorous, and narratively coherent outputs in scientific applications.

\clearpage
\clearpage

\clearpage
\clearpage

\clearpage
\section{QK-Pair Examples}
\label{app:qk_examples}
\begin{figure*}[hb]
\begin{tcolorbox}[title=\mt~Example,
  left=1mm, right=1mm, top=0mm, bottom=0mm, colback=white,
  sharp corners, boxrule=0.3mm]

{\color{spblue}\textbf{Type:} macro mechanics} \\
{\color{spred}\textbf{Question:}} Explain the concept of fatigue life prediction using the S-N curve. \\
{\color{spgreen}\textbf{Key:}} An S-N curve (stress vs. number of cycles) plots fatigue strength over cyclic life. It helps predict the number of cycles a material can withstand at a given stress amplitude before failure.

\par\vspace{2mm}

{\color{spblue}\textbf{Type:} micro mechanics} \\
{\color{spred}\textbf{Question:}} What is spinodal decomposition? \\
{\color{spgreen}\textbf{Key:}} Spinodal decomposition is a phase separation mechanism in the unstable region of a free energy curve where \( \frac{\partial^2 G}{\partial c^2} < 0 \), and small composition fluctuations grow spontaneously without a nucleation barrier.

\par\vspace{2mm}

{\color{spblue}\textbf{Type:} electrical and electronic} \\
{\color{spred}\textbf{Question:}} Why are ceramics brittle and metals ductile? \\
{\color{spgreen}\textbf{Key:}} Ceramics have strong directional ionic/covalent bonds with limited slip systems \( \rightarrow \text{ brittle} \). Metals have metallic bonds with abundant slip systems \( \rightarrow \text{ ductile} \). This difference in plasticity is key to fracture behavior.

\par\vspace{2mm}

{\color{spblue}\textbf{Type:} optical and microscopic} \\
{\color{spred}\textbf{Question:}} Draw and explain the Ewald sphere. \\
{\color{spgreen}\textbf{Key:}} The Ewald sphere is a sphere in reciprocal space of radius \( \frac{1}{\lambda} \). A reciprocal lattice point lying on this sphere satisfies Bragg’s law, leading to a diffraction spot.

\par\vspace{2mm}

{\color{spblue}\textbf{Type:} thermodynamical} \\
{\color{spred}\textbf{Question:}} What is the difference if you put a hot metal into dry ice and an ice-water mixture? \\
{\color{spgreen}\textbf{Key:}} Dry ice is around \( -78^\circ\text{C} \), while an ice-water mixture remains at \( 0^\circ\text{C} \). A hot metal will cool more rapidly in contact with dry ice than in an ice-water mixture.

\par\vspace{2mm}

{\color{spblue}\textbf{Type:} semiconductor} \\
{\color{spred}\textbf{Question:}} Describe the thermal conductivity of electrons in a metal. \\
{\color{spgreen}\textbf{Key:}} In metals, thermal conductivity of electrons is high because electrons are free to transport energy. According to the Wiedemann–Franz law, the ratio of thermal to electrical conductivity is proportional to temperature.

\par\vspace{2mm}

{\color{spblue}\textbf{Type:} math} \\
{\color{spred}\textbf{Question:}} Gary likes to walk around the edge of the local park, which is a rectangle that measures 1.5 miles by 6 miles. If he walks at 3 miles/hour, how many hours does he spend walking? \\
{\color{spgreen}\textbf{Key:}} Length of the two long sides: \( 6 \times 2 = 12\ \text{miles} \). Length of the two short sides: \( 1.5 \times 2 = 3\ \text{miles} \). Total perimeter: \( 12 + 3 = 15\ \text{miles} \). Time: \( \frac{15\ \text{miles}}{3\ \text{miles/hour}} = 5\ \text{hours} \). Answer: \( \boxed{5} \)

\par\vspace{2mm}

{\color{spblue}\textbf{Type:} geoscience} \\
{\color{spred}\textbf{Question:}} Which of the following is the best example of a multilingual country? [A. Japan / B. The United States / C. Brazil / D. Canada / E. Iceland] \\
{\color{spgreen}\textbf{Key:}} D

\par\vspace{2mm}

{\color{spblue}\textbf{Type:} biology} \\
{\color{spred}\textbf{Question:}} Can transcranial direct current stimulation be useful in differentiating unresponsive wakefulness syndrome from minimally conscious state patients? \\
{\color{spgreen}\textbf{Key:}} Yes, tDCS could be useful in identifying residual connectivity markers in clinically-defined UWS, who may lack purposeful behavior as a result of a motor-output failure.

\end{tcolorbox}
\vspace{-3.5mm}
\caption{Example \textit{(Q,K)} of \mt.}
\label{fig:multitask_example}
\end{figure*}

\begin{figure*}[h]
\begin{tcolorbox}[title=\st~Example,
  left=1mm, right=1mm, top=0mm, bottom=0mm, colback=white,
  sharp corners, boxrule=0.3mm]

{\color{spblue}\textbf{Background}} \\
{\color{spred}\textbf{Question (Task Description):}} The rapid advancement in materials science demands innovative methods to discover new materials with exceptional properties. Traditional experimentation is time-consuming and costly, motivating the integration of machine learning to predict material behaviors and identify novel compounds efficiently. \\
{\color{spgreen}\textbf{Key:}} $\varnothing$

\par\vspace{2mm}

{\color{spblue}\textbf{Methodology}} \\
{\color{spred}\textbf{Question (Task Description):}} We propose using deep learning models, such as graph neural networks, to analyze large datasets of material properties. These models will enable the prediction of atomic interactions and facilitate the identification of promising new materials with desired characteristics.\\
{\color{spgreen}\textbf{Key:}} $\varnothing$

\par\vspace{2mm}

{\color{spblue}\textbf{Impact}} \\
{\color{spred}\textbf{Question (Task Description):}} This research aims to accelerate materials discovery, offering substantial contributions to academic understanding and providing industry with tools to develop advanced materials for sectors like energy, electronics, and biotechnology.\\
{\color{spgreen}\textbf{Key:}} $\varnothing$

\end{tcolorbox}
\vspace{-3.5mm}
\caption{Example \textit{(Q,$\varnothing$)} pair of \st.}
\label{fig:singletask_example}
\end{figure*}

\end{document}